\documentclass[letterpaper]{article} 
\usepackage[preprint]{aaai2027}  
\usepackage[hyphens]{url}  
\usepackage{graphicx} 
\usepackage{natbib}  
\usepackage{caption} 
\usepackage{algorithm}
\usepackage{algorithmic}
\usepackage{todonotes}
\usepackage{xcolor}

\usepackage{newfloat}
\usepackage{listings}
\DeclareCaptionStyle{ruled}{labelfont=normalfont,labelsep=colon,strut=off} 
\floatstyle{ruled}
\newfloat{listing}{tb}{lst}{}
\floatname{listing}{Listing}

\usepackage{booktabs}

\title{\textsc{OmouAI}: Argumentative Human-AI Policy Deliberation \\ with Simulated Personas}
\author{
    Stylianos Loukas Vasileiou\textsuperscript{\rm 1},
    Antonio Rago\textsuperscript{\rm 2},
    William Yeoh\textsuperscript{\rm 3} and
    Georgina Curto\textsuperscript{\rm 4}
}
\affiliations{
    \textsuperscript{\rm 1}New Mexico State University,
    \textsuperscript{\rm 2}King's College London,
    \textsuperscript{\rm 3}Washington University in St. Louis, 
    \textsuperscript{\rm 4}United Nations University

}

\begin{document}

\maketitle

\begin{abstract}
Debates amongst agents driven by large language models (LLMs) have demonstrated vast potential in various applications, but when these interactions include humans and take place in high-stakes environments, e.g., in public policy deliberations, they are beset with issues such as sycophancy and a lack of faithful explanations. To tackle these issues, we present \textsc{OmouAI}, an interactive and inclusive deliberation system that uses LLMs in combination with computational argumentation, a field which excels in representing and reasoning within debates. \textsc{OmouAI} allows a human user to deliberate policy claims for real-world challenges with simulated personas, e.g., representing stakeholders, domain experts or devil's advocates, towards reducing sycophancy. Each persona generates its own arguments, and the arguments of all parties form a shared argumentation framework. Users can then contest, add and revise arguments, providing crucial human oversight. Then, arguments are evaluated using deterministic argumentative semantics against external goals, such as the UN Sustainable Development Goals, guaranteeing faithful explanations. The advancement or worsening of the goals thus serve as indicators for the policy recommendations. 
\end{abstract}

\section{Introduction}


AI agents built on large language models (LLMs) now support decisions in high-stake domains such as medicine, law, and public administration \citep{wang2024survey}. 
These agents have been shown to be particularly powerful when their decision making is the result of debate \cite{Khan_24}. 
Indeed, the Habermas Machine \citep{tessler2024habermas} showed that LLMs can help citizens find common ground on divisive policy questions. Policy deliberation is a natural setting for such agents, since policy questions are contested by parties who hold different values and weigh the same evidence differently. These agents, however, often generate fluent text and reason over unstructured inputs without clearly exposing their reasoning, so their outputs cannot be faithfully explained or reliably contested. In high-stakes deliberations, reasoning that cannot be explained, evaluated or contested is of limited use. Further, LLMs are known to display ``sycophantic'' behaviour towards their users \cite{Bo_26}, which has no place in effective deliberation.


Meanwhile, computational argumentation \cite{Dung_95} is an area 
which has shown great suitability for representing and reasoning with 
debates, whether between machines \cite{Gorur_26}, humans \cite{Lawrence_23,Ruiz-Dolz_23,Irwin_22,Goffredo_25} or a combination thereof \cite{Rago_23}
.  
Recently, it has been proposed that argumentative mechanisms 
facilitate the contestation of LLM recommendations through formal frameworks for transparent, verifiable reasoning with well-defined semantics \citep{vasileiou2026argumentative}. 
Indeed, recent works 
\citep{freedman2025argumentative,zhu2025argrag,dejl2026argllmapp} allow for LLMs to generate and weigh arguments for and against a claim, then compute the verdict with gradual semantics over the resulting argumentation framework.
However, such systems have not been built specifically for public policy deliberation. In turn, deliberation systems address the opposite problem: the Habermas Machine and related systems \citep{tessler2024habermas,ma2025humanai} aggregate many human views into a statement, but they treat LLMs as ``black box'' mediators, leaving the underlying reasoning, and thus the justifications for decisions, unexposed. 

\begin{figure*}[t!]
\centering
\includegraphics[width=\textwidth]{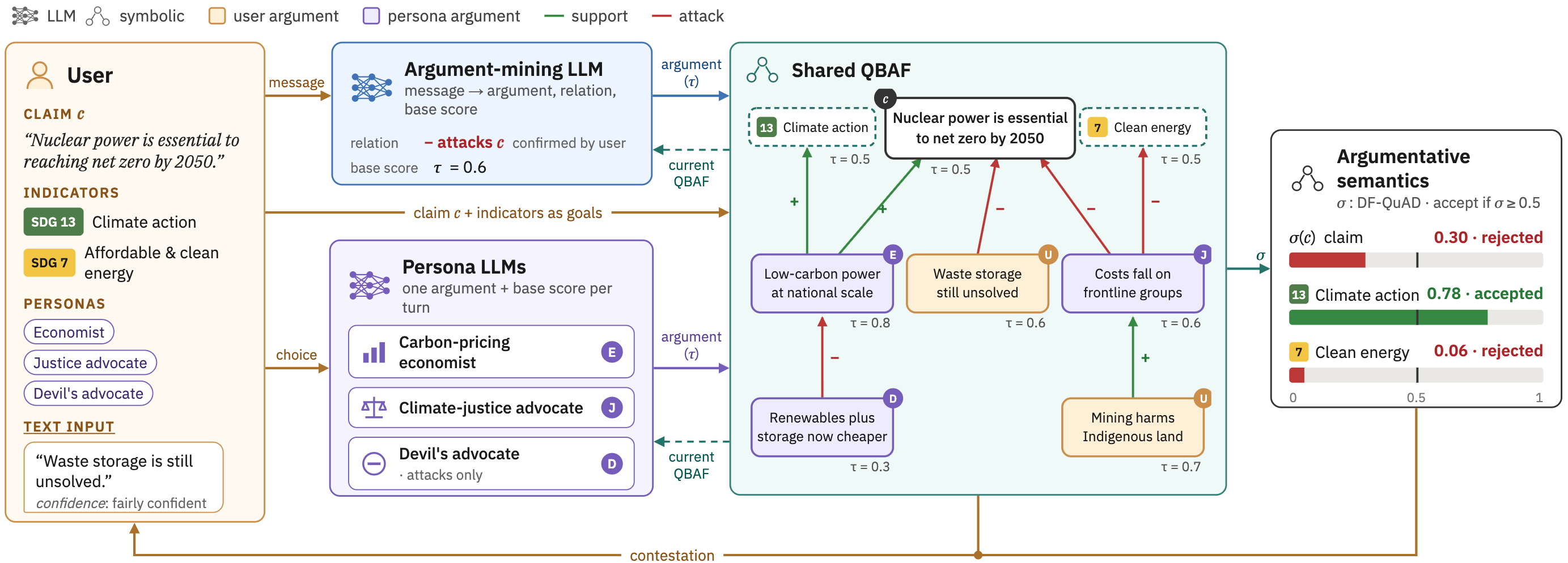}
\caption{The \textsc{OmouAI} pipeline for a climate change policy example.}
\label{fig:pipeline}
\end{figure*}

We present \textsc{OmouAI},\footnote{\textsc{OmouAI} is derived from the word \textit{omou}, which means \textit{deliberating} in Japanese and \textit{together} in Greek. Taken together, \textsc{OmouAI} means \textit{deliberating together with AI}.} to the best of our knowledge the first system for collaborative human-AI decision-making which combines LLMs and argumentation specifically for public policy deliberation. \textsc{OmouAI} allows for: (i) an inclusive deliberative system for real-world challenges, where the human discusses with a diversity of personas (stakeholders, domain experts, or devil's advocates); (ii) a joint, explainable and contestable human-agent argumentative thread, where arguments are generated and scored based on the personas' specifications; and (iii) impact evaluation of the proposed arguments towards real-world goals, e.g., the United Nations Sustainable Development Goals.

\section{System Overview}

\textsc{OmouAI} 
is a neurosymbolic system comprising an LLM layer and an argumentation layer. Before a deliberation starts, the user enters a policy claim, chooses the personas that take part, and selects indicators representing goals against which the deliberation is to be measured. During the deliberation, the LLM layer generates arguments ($\alpha_p$) and scores them with a base score ($\tau$), i.e., an intrinsic strength, on behalf of each persona, and mines the user's messages into arguments ($\alpha_u$) and base scores. In the argumentation layer, these two sets of arguments are combined in a shared quantitative bipolar argumentation framework (QBAF) \citep{baroni2018many} with arguments representing the claim $c$ and each selected indicator ($\alpha_i$), all of which are evaluated with a gradual semantics ($\sigma$). Figure~\ref{fig:pipeline} shows the pipeline.

\noindent \textbf{QBAFs.}~The shared QBAF is a tuple $\langle \mathcal{A}, \mathcal{R}^-, \mathcal{R}^+, \tau\rangle$ comprising a set of arguments $\mathcal{A} = \{c\} \cup \alpha_p \cup \alpha_u \cup \alpha_i$, disjoint attack and support relations $\mathcal{R}^-, \mathcal{R}^+ \subseteq \mathcal{A}\times\mathcal{A}$, and a base score function $\tau : \mathcal{A}\to[0,1]$. A gradual semantics $\sigma : \mathcal{A}\to[0,1]$ assigns each argument a strength computed from its base score and the strengths of its attackers and supporters. \textsc{OmouAI} implements several such semantics, including DF-QuAD \citep{rago2016discontinuity} and the quadratic energy model \citep{potyka2018continuous}, and the user can switch between them at runtime. The QBAF is a merged set of trees rooted at the policy claim $c$ and the indicators. 
Every argument in \textsc{OmouAI} additionally carries a type, which is empirical for a statement about the present state of affairs, predictive for a statement about what the policy would cause, or normative for a statement about what ought to be done
.

\noindent \textbf{Personas.}~A persona is a specification of three to five value commitments, a specification stating the kinds of evidence the persona accepts, and an epistemic disposition stating how cautiously it reasons from that evidence (following value-belief-norm theory \citep{stern2000toward,long2025chain}). These specifications are the only persona-specific content given to the LLM, 
which are minimised since persona prompting degrades when irrelevant attributes are added \citep{park2024generative,hu2024quantifying}. The user selects any number of personas to take part in the deliberation. A persona can be a stakeholder, which represents an affected party and whose values condition both generation and scoring; a domain expert, which argues only from evidence; or a devil's advocate, which only attacks and serves to test arguments. 

\noindent \textbf{Argument generation and scoring ($\alpha_p$).}
Given a target argument and the arguments already attached to it, $\alpha_p$ returns a single new attacker or supporter of the target on behalf of the persona
, or declines if the persona has nothing to add.
$\tau$ then assigns the base score according to the argument's type, scoring an empirical or predictive argument by the evidential support the persona finds for it and a normative argument by the degree to which the persona's commitments endorse it. 
The claim and indicator arguments start with a base score of $0.5$, indicating a neutral stance.

\noindent \textbf{Argument mining ($\alpha_u$).}
The user's own contributions are provided as free-text messages, and each message is first classified (by an LLM) as an argument, a request or a question \citep{chen2024exploring}. An argument is a message that takes a position on the claim or on an existing argument. For such a message, \textsc{OmouAI} returns a restatement as a single argument, and a proposed attack or support relation to an existing argument, and the user confirms or redirects the proposed relation before the argument enters the QBAF. The user's argument is also assigned a base score $\tau$ based on the confidence the user stated for it on a five-level scale. A request is a message that asks a persona for an argument, which results in a new argument in $\alpha_p$. A question is a message that asks about the current state of the deliberation, and it is answered from the computed record of the argumentation layer.

\noindent \textbf{Evaluation ($\sigma$).}
The argumentation layer computes 
a strength for all arguments, including those representing the claim and indicators,
and thus the deliberation results in a verdict for each goal alongside the verdict on the claim. 
The implemented semantics satisfy the base score and relation contestability properties in \citep{freedman2025argumentative}, which guarantee intuitive behaviour such as
a user's intervention always moving the verdict in the direction the user intends.


\noindent \textbf{Responsible Use.} \textsc{OmouAI} is a human-AI argumentation-support tool, not a prescriptive decision-making system. The generated arguments reflect the biases of the underlying models and the value systems of the represented stakeholders. Domain experts must evaluate the outputs before these are considered for real-world decision making purposes. 


\bibliography{references}


\end{document}